\documentclass{esannV2}
\usepackage[dvips]{graphicx}
\usepackage[latin1]{inputenc}
\usepackage{amssymb,amsmath,array}
\usepackage{url}
\graphicspath{ {./} }
\newlength{\bibitemsep}
\newlength{\bibparskip}
\let\oldthebibliography\thebibliography
\renewcommand\thebibliography[1]{%
  \oldthebibliography{#1}%
  \setlength{\parskip}{\bibitemsep}%
  \setlength{\itemsep}{\bibparskip}%
}
\usepackage{capt-of}
\usepackage[dvipsnames]{xcolor}
\usepackage{placeins}
\usepackage{hhline,colortbl}
\usepackage{float}
\usepackage{eso-pic}
\usepackage{array}
\usepackage{booktabs}
\usepackage{mathtools}
\newcommand{\f}[1]{\underline{\textbf{#1}}}
\newcommand{\s}[1]{\textbf{#1}}

\AtBeginDocument{\AddToShipoutPictureFG*{\AtTextUpperLeft{\put(0,\LenToUnit{40pt}){\parbox{\textwidth}{\centering\bfseries
					29th European Symposium on Artificial Neural Networks, Computational Intelligence and Machine Learning (ESANN), Bruges, Belgium, 2021
}}}}}

\begin{document}
%style file for ESANN manuscripts
\title{Semantic Prediction:\\ Which One Should Come First,\\ Recognition or Prediction?}

%***********************************************************************
% AUTHORS INFORMATION AREA
%***********************************************************************
\author{Hafez Farazi, Jan Nogga, and Sven Behnke
\vspace{.3cm}\\
\textit{University of Bonn, Computer Science Institute VI, Autonomous Intelligent Systems} \\
Friedrich-Hirzebruch-Allee 5, 53115 Bonn, Germany  \\
\{farazi, nogga, behnke\}@ais.uni-bonn.de
}

%***********************************************************************
% END OF AUTHORS INFORMATION AREA
%***********************************************************************
\vspace{-10px}
\maketitle

\begin{abstract} The ultimate goal of video prediction is not forecasting future pixel-values given some previous frames. Rather, the end goal of video prediction is to discover valuable internal representations from the vast amount of available unlabeled video data in a self-supervised fashion for downstream tasks. One of the primary downstream tasks is interpreting the scene's semantic composition and using it for decision-making. For example, by predicting human movements, an observer can anticipate human activities and collaborate in a shared workspace. There are two main ways to achieve the same outcome, given a pre-trained video prediction and pre-trained semantic extraction model; one can first apply predictions and then extract semantics or first extract semantics and then predict. We investigate these configurations using the Local Frequency Domain Transformer Network (LFDTN) as the video prediction model and U-Net as the semantic extraction model on synthetic and real datasets.
\end{abstract}
\vspace{-5px}
\section{Introduction}
\vspace{-5px}
\begin{figure}[t]
\vspace{-40px}
  \centering
    \includegraphics[width=0.75\linewidth]{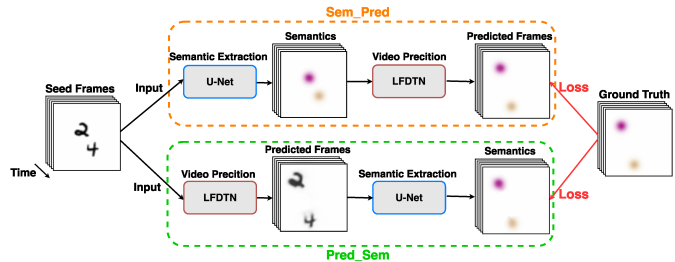}
    \vspace{-10px}
    \caption{Two possible configurations for semantic prediction. We want to predict future semantics using a pre-trained video prediction (LFDTN) and a pre-trained semantic extraction model (U-Net).}
    \label{thumbnail}
\vspace{-15px}
\end{figure}

The task of forecasting subsequent frames of a video sequence given several seed frames is commonly referred to as video prediction. As an accurate model must account for the dynamics of the observed scene, it is evident that the video prediction system must develop an understanding thereof. For this reason, video prediction is often posed as a pretext task for self-supervised scene comprehension. In our previous work, we introduced LFDTN~\cite{LFDTN2021}, an approach for video prediction and motion segmentation that is fully interpretable and lightweight.
Employing self-supervised learning, the representations formed by our model describe the pixel-to-pixel transformation between frames. %Additionally, our model can provide a semantic map of each frame into dynamic and static components.
Using this technique, we outperformed other video prediction models on real-world video data from traffic cameras in a fully interpretable manner while utilizing only unlabeled data.

In this work, we investigate the capability of LFDTN to integrate with existing and preferably pre-trained recognition systems. In our specific use case, the ultimate goal is to anticipate human poses in a shared human-robot workspace. To achieve this, we developed two separate pipelines; one for predicting video frames~\cite{LFDTN2021}~\cite{farazi2020} and the other for extracting human poses from still frames~\cite{pose2021}. 
Since the end goal is predicting semantics in future frames and not necessarily photo-realistic video predictions, we can reach this result in two different configurations as illustrated in Fig.~\ref{thumbnail}. On the one hand, it might be reasonable to predict future frames and then extract semantics from the predictions. On the other hand, it is also plausible to first obtain semantics from the seed frames and then apply the video prediction pipeline to the resulting semantic channels. In either case it is also necessary to fine-tune the combined system.

In this work, the LFDTN model is used for the video prediction task because we previously showed that LFDTN outperforms other baseline models on the synthetically produced ``Moving MNIST ++" dataset and real traffic scene videos~\cite{LFDTN2021}. In this work, we also employ the well-known and widely used U-Net~\cite{ronneberger2015unet} model to extract semantics. The code and dataset of this paper are publicly available.\footnote{ \url{https://github.com/AIS-Bonn/Pred_Semantic}.}
\vspace{-5px}
\section{Related Work}
\vspace{-5px}
While numerous approaches to video prediction have been proposed, the most effective ones utilize some form of deep learning. Most of them employ recurrent layers like LSTMs or GRUs at their models' core which are used as black-boxes. While requiring a large number of parameters, they lack interpretability. Two successful examples are Location-Dependent Video Ladder Networks~\cite{AziziFar2018} and PredRNN++~\cite{Wang2018PredRNNTA}, which employs a stack of LSTM modules to predict plausible future frames. 

Wagner et al.~\cite{Wagner2017}, proposed a semantic prediction model based on pre-trained feedforward networks to extract semantics followed by Conv-LSTM to predict future semantics. Bei et al.~\cite{bei2021learning} recently showed that conditioning video prediction on semantics and optical flow yields better results. Luc et al.~\cite{luc2017predicting} investigate the problem of future pixel-wise semantic segmentation prediction. Similar to our work, they examined different possible configurations for the segmentation-prediction problem. In contrast to their work, we are investigating the issue with multiple overlapping semantic channels simultaneously. Besides, we are ultimately interested in easier-to-predict blob-like human joints and not fine-grained pixel-wise semantic segmentation. Additionally, they utilized a deep model with a massive number of parameters compared to LFDTN, which is interpretable and lightweight. Another difference is that they did not explore the end-to-end refinement of the joint segmentation-prediction model; instead, they applied the segmentation model as a given fixed network.

\vspace{-5px}
\section{Models}
\vspace{-5px}
In this section, we briefly introduce the components used in our experiments.

\noindent\textbf{Video Prediction (LFDTN):} The core functionality provided by Local Frequency Domain Transformer Networks is the capability to describe changes in an observed image as a collection of local linear transformations, transport inferred shifts into the future, and consequently apply them to produce a prediction of the next frame's content. The first part of these three distinct tasks is handled by a process which can be described as a \textit{Local Fourier Transform} (LFT), a Fourier-related transform resembling the STFT for a 2D signal. For a given frame $x_{t}$, overlapping tiles are extracted and windowed by a function $w$ to produce $x_{t,u,v}$, a collection of tapered windows on $x_{t}$ around the image coordinates $\{u,v\}$. For each, the FFT $\mathcal{X}_{t,u,v}$ is calculated. Given $\mathcal{X}_{t-1,u,v}$ and $\mathcal{X}_{t,u,v}$ (the LFTs of two consecutive frames $x_{t-1}$ and $x_t$) the \textit{local phase difference} is then defined element-wise as:
\vspace{-5px}
\begin{equation}
\mathcal{PD}_{t-1,u,v} \coloneqq \frac{\mathcal{X}_{t,u,v}\overline{\mathcal{X}_{t-1,u,v}}}{|\mathcal{X}_{t,u,v}\overline{\mathcal{X}_{t-1,u,v}}|}.
\end{equation}
\vspace{-5px}

The local phase differences encode the image shift observed around $\{u,v\}$ and serve as a content-independent description of the local image transformation. Since local adversities sometimes perturb the phase differences in addition to the fact that the shifts are generally not spatiotemporally constant, a lightweight learnable convolutional transform network $\mathcal{MM}$ filters and transports them one time-step ahead as:
\vspace{-5px}
\begin{equation}
\widehat{\mathcal{PD}}_{t,u,v} = \mathcal{MM}(\mathcal{PD}_{t-1,u,v}).
\end{equation}
\vspace{-5px}

It should be noted that $\mathcal{MM}$ is designed with an intentional bottleneck which enforces the representation of $\widehat{\mathcal{PD}}_{t,u,v}$ to be a vector field at the output layer, which is readily accessed and can explain the final prediction results well. Some samples of these vector fields are shown in Fig.~\ref{result}.

Next, a prediction of the local views on $x_{t+1}$ is formed via the \textit{local phase addition} given by:
\vspace{-5px}
\begin{equation}
    \widehat{\mathcal{X}}_{t+1,u,v} = \mathcal{X}_{t,u,v} \cdot \widehat{\mathcal{PD}}_{t,u,v}
\label{loc_view}
\end{equation}
\vspace{-5px}

and subsequently obtaining their inverse Fourier transforms $\hat{x}_{t,u,v}$. Additionally, the effect of the local shifts on the tapering windows are accounted for by repeating this step for the Fourier transform of the window function $\mathcal{W}$ yielding:
\vspace{-5px}
\begin{equation}
\hat{w}_{t,u,v} \coloneqq \operatorname{iFFT}(\operatorname{phase\_add}(\mathcal{W}, \widehat{\mathcal{PD}}_{t,u,v})).
\end{equation}
\vspace{-10px}

Using both, the next video frame $x_{t+1}$ is reconstructed as:
\vspace{-5px}
\begin{equation}\label{eq:OLA}
x_{t+1}[n,m] = \frac{\sum\limits_{u=-\infty}^\infty \sum\limits_{v=-\infty}^{\infty} \hat{x}_{t+1,u,v}[n,m]\hat{w}[n-uH, m-vH]}{\sum\limits_{u=-\infty}^\infty \sum\limits_{v=-\infty}^{\infty} \hat{w}^{2}[n-uH, m-vH]} ,
\end{equation}
\vspace{-10px}

which represents the inverse local Fourier Transform given shifted windows. Alternatively, the whole process can be understood as the extension of our previous Frequency Domain Transformer Networks~\cite{farazi2019} to local shifts via LFT. The sequence of analysis, then transport and predict, and finally, synthesis described above can also be applied channel-wise. This means that any spatial signal, for example segmented video, also constitutes a valid input.

\noindent\textbf{Semantic Extraction (U-Net):} U-Net~\cite{ronneberger2015unet} is a well-known architecture originally developed for medical image segmentation tasks. It consists of a convolutional encoder network followed by a decoder network. %The input at each layer of the decoder network is concatenated along the channel dimension with an appropriately cropped output feature map from the encoder network at the same level. This represents a comparatively high-resolution input to the decoder layer, as the encoder typically downsamples at each layer while increasing the number of extracted features. In this manner, U-Net can form rich representations of the input data but is at the same time relieved of the effort to reconstruct redundancies in the decoder representations, which culminates in the output of a spatially detailed segmentation map at its top level. 
Because we have overlapping digits in our synthetic dataset, we do not employ softmax and instead train the unnormalized responses using MSE loss.
\vspace{-8px}
\section{Experimental Results}
\vspace{-8px}
\begin{figure}[t]
  \vspace{-50px}
  \centering
    \includegraphics[width=0.95\linewidth]{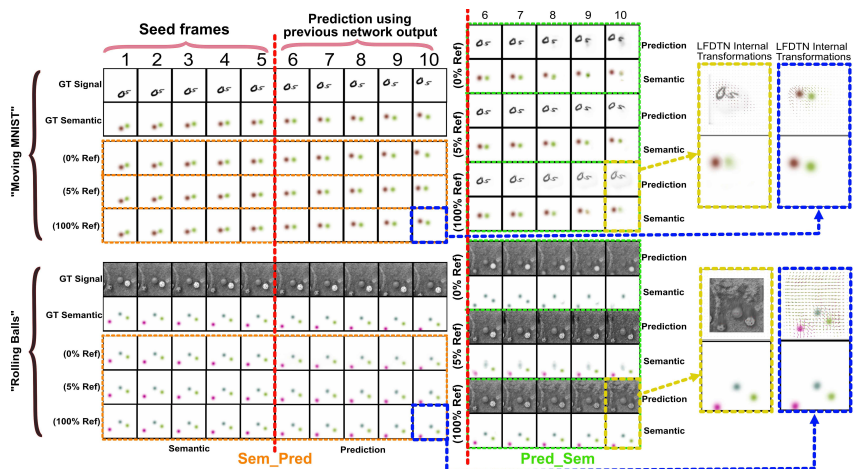}
    \vspace{-15px}
    \caption{Sample results on two possible configurations for semantic prediction. Note different blobs' colors represent distinct semantic channels. For the fully fine-tuned models, the internally formed transformation of LFDTN are shown.}
    \label{result}
    \vspace{-15px}
\end{figure}
\noindent\textbf{Dataset:} We used two datasets for our experiments. We adapted the synthetic ``Moving MNIST" dataset and recorded some videos containing real moving soccer balls which we call the ``Rolling Balls" dataset. Two distinct MNIST digits are randomly placed into a $64\times64$ canvas with random velocities in the synthetic dataset. There are six visually different soccer balls rolling in a $128\times128$ frame in the real dataset. We also create an accompanying sequence indicating the semantics. For this purpose, we make $K$ channels per frame featuring a Gaussian activity blob at the center of mass of each present digit or soccer ball in the corresponding channel. Note that $K$ is ten in the ``Moving MNIST" dataset and six in the ball dataset. We consider these datasets suitable because both U-Net-like models, for the recognition model, and LFDTN, the prediction model, were previously separately tested on similar data. One sample sequence of each dataset is shown in Fig.~\ref{result}.

% \noindent\textbf{Dataset:} Since our experiments require ground truth for the semantic extraction task and the joint semantic prediction task, we decided to adopt the ``Moving MNIST" dataset. Two distinct MNIST digits are randomly placed into a $64\times64$ canvas with random velocities. To synthesize subpixel movements, the digits are moved in higher resolution canvases and downsampled afterwards. We also create an accompanying sequence indicating the semantics. For this purpose, we create ten channels per frame featuring a Gaussian activity blob at the center of mass of each present digit in the corresponding channel. We consider this dataset suitable because both U-Net, the recognition model, and LFDTN, the prediction model, were previously separately verified on such data. One sample sequence is shown in Fig.~\ref{result}.

\noindent\textbf{Pipelines and Training:} Each dataset is used to train two pipelines, each corresponding to one semantics-prediction configuration: \textit{Sem\_Pred} and \textit{Pred\_Sem}. In the \textit{Sem\_Pred} pipeline, the semantics of the seed frames are first extracted using a pre-trained U-Net, producing $K$ channels, one channel for each object class, with an activity blob in the corresponding semantic channel. The output is then passed to the LFDTN, which predicts the future activity blobs per channel. Finally, the result is compared to the ground truth semantic prediction. The semantic model has been trained to recognize multiple objects (MNIST digits and soccer balls) in single images and generate blob-like responses in corresponding channels. In contrast, the LFDTN has been pre-trained on freely moving Gaussian blobs.

The opposite order of modules is applied in the \textit{Pred\_Sem} pipeline. In the first step, the LFDTN observes the seed frames and outputs a prediction of the future frames in the signal level. These predicted frames are then parsed by U-Net, yielding activity blobs placed at future object positions. This output is, as before, compared to the ground truth semantic predictions. While we used the same pre-trained U-Net, the LFDTN in this pipeline was trained on sequences of frames in the signal level. 

Each configuration's experiment is repeated by fine-tuning on $\{1, 5, 10, 20, 100\}$ percent of the entire training set, respectively. End-to-end refinement is possible, because both U-Net and LFDTN are fully differentiable. In all experiments, all models use the same number of trainable parameters. For the \textit{Sem\_Pred} trials, this means that the LFDTN shares its weights across all input class channels, which is necessary to ensure that the experiments are fair.

The pre-trained models were trained on approximately two-fold more data samples to simulate the real-world scenarios in which acquiring unlabeled videos and annotated single frame semantics is much simpler than annotating videos. We also trained both configurations end-to-end without intermediate loss, from scratch on the smaller joint dataset, which is denoted by star \textit{(*)} in Table~\ref{Table_results1}. Note that for the synthetic experiments, U-Net holds $504k$ learnable parameters while the LFDTN has $26k$ parameters. For the experiments on real data, U-Net contains $732k$ learnable parameters, and the LFDTN has $67k$ parameters. Overall, these experiments provide a guideline to find the right balance between pre-training and end-to-end fine-tuning in similar scenarios.

\noindent\textbf{Results:} As shown in Table~\ref{Table_results1}, it is better to first extract semantics and then perform video prediction. We also observe diminishing returns when fine-tuning on larger fractions of the dataset, which is encouraging since acquiring semantics from the video is very costly, and one needs to find the right balance between the model performance and the amount of semantically annotated videos. The results are visually plausible in terms of the predicted positions of the class activity blobs, which is reflected by low errors. Sample results for some of the experiments are depicted in Fig.~\ref{result}. We observe that at least a bit end-to-end fine-tuning is necessary since pre-trained models were accustomed to the sharp and perfect input signal, which struggles after receiving the output of the other model. It is evident from the result that even if one has full video annotations, starting with pre-trained models will still be beneficial. Our research confirms the optimum configuration suggested by Bei et al.~\cite{bei2021learning} on our totally different models and datasets. We hypothesize that extracting semantics at the first step would ease the burden on the following model since getting reliable predictions requires some understanding of semantics.

\begin{table}[t]
\vspace{-52px}
\begin{center}
	\scriptsize
	\begin{tabular}
		{l || c | c | c|| c| c| c||}
		
		\scriptsize{\hspace{0px}}&  
		\multicolumn{3}{c||}{\scriptsize{\hspace{0px}``Moving MNIST"\hspace{0px}}} &
		\multicolumn{3}{c||}{\scriptsize{\hspace{0px}``Rolling Balls"\hspace{0px}}}\\
		
		\scriptsize{Model\hspace{0px}} & 
		\scriptsize{\hspace{0px}L1\hspace{0px}} & 
		\scriptsize{\hspace{0px}MSE\hspace{0px}} &
		\scriptsize{\hspace{0px}DSSIM\hspace{0px}} & 
		\scriptsize{\hspace{0px}L1\hspace{0px}} & 
		\scriptsize{\hspace{0px}MSE\hspace{0px}}  &
		\scriptsize{\hspace{0px}DSSIM\hspace{0px}}\\

        \arrayrulecolor{BurntOrange}
		\hline
		\hline  % L1   L2  SSIM       #

		\scriptsize{Sem\_Pred (*)}       &\s{0.0014}&\s{0.00028}&   0.0039    &   0.00100  &   0.000223  &   0.0026 \\
		\scriptsize{Sem\_Pred (0\%)}     &   0.0033 &   0.00138 &   0.0089    &   0.00111  &   0.000211  &   0.0037 \\
		\scriptsize{Sem\_Pred (1\%)}     &   0.0024 &   0.00084 &   0.0062    &\f{0.00082} &   0.000198  &\f{0.0019}\\
		\scriptsize{Sem\_Pred (5\%)}     &   0.0017 &   0.00049 &   0.0042    &\s{0.00085} &   0.000197  &   0.0020 \\
		\scriptsize{Sem\_Pred (10\%)}    &   0.0016 &   0.00049 &   0.0041    &   0.00088  &\s{0.000196} &   0.0020 \\
		\scriptsize{Sem\_Pred (20\%)}    &   0.0014 &   0.00040 &\s{0.0035}   &   0.00088  &   0.000200  &   0.0021 \\
		\scriptsize{Sem\_Pred (100\%)}   &\f{0.0011}&\f{0.00025}&\f{0.0024}   &   0.00088  &\f{0.000194} &\s{0.0020} \\	
		\arrayrulecolor{green}
		\hline
		\hline
		\scriptsize{Pred\_Sem (*)}       &   0.0034 &   0.00103 &   0.0181    &   0.00093  &   0.000297  &   0.0028 \\
		\scriptsize{Pred\_Sem (0\%)}     &   0.0044 &   0.00148 &   0.0212    &   0.00150  &   0.000702  &   0.0040 \\
		\scriptsize{Pred\_Sem (1\%)}     &   0.0034 &   0.00111 &   0.0162    &   0.00101  &   0.000354  &   0.0030 \\
		\scriptsize{Pred\_Sem (5\%)}     &   0.0030 &   0.00096 &   0.0136    &   0.00099  &   0.000377  &   0.0027 \\
		\scriptsize{Pred\_Sem (10\%)}    &   0.0029 &   0.00094 &   0.0127    &   0.00104  &   0.000343  &   0.0034 \\
		\scriptsize{Pred\_Sem (20\%)}    &   0.0030 &   0.00095 &   0.0140    &   0.00086  &   0.000261  &   0.0026 \\
		\scriptsize{Pred\_Sem (100\%)}   &   0.0028 &   0.00080 &   0.0133    &   0.00090  &   0.000336  &   0.0022 \\	
		
\end{tabular}
\vspace{-5px}
\footnotesize{
\caption{Testset results for ``Moving MNIST" and ``Rolling  Balls" datasets.}}
\label{Table_results1}
\end{center}
\vspace{-25px}
\end{table}

\vspace{-6px}
\section{Conclusion}
 
\vspace{-6px}
Our experimental result indicates that first performing object recognition and then applying prediction yields better semantic predictions. We also showed that fine-tuning the joint pipeline provides diminishing returns, indicating that some co-adaption of the models is beneficial. Yet, considering the high cost of annotating videos and depending on the application, it is not necessary to train the full model end-to-end from scratch. Starting with pre-trained individual components followed by a bit of joint model fine-tuning is sufficient. Overall, we showed that combining the pre-trained U-Net and LFDTN models works very well. We will build on these findings for future works and examine the human pose prediction using \textit{Sem\_Pred} with little end-to-end refinements next.

\vspace{-6px}
{\footnotesize \paragraph{Acknowledgment:}
\label{acknowledgment}
This work was funded by grant BE 2556/16-2 (Research Unit FOR 2535 Anticipating Human Behavior) of the German Research Foundation (DFG).
}
\vspace{-6px}
\bibliographystyle{unsrt}
{\small \bibliography{esannV2}}

\begin{thebibliography}{10}

\bibitem{LFDTN2021}
H.~Farazi, J.~Nogga, and S.~Behnke.
\newblock Local frequency domain transformer networks for video prediction.
\newblock In {\em IJCNN}, 2021.

\bibitem{farazi2020}
H.~Farazi and S.~Behnke.
\newblock Motion segmentation using frequency domain transformer networks.
\newblock In {\em ESANN}, 2020.

\bibitem{pose2021}
S.~Bultmann and S.~Behnke.
\newblock Real-time multi-view {3D} human pose estimation using semantic
  feedback to smart edge sensors.
\newblock In {\em RSS}, 2021.

\bibitem{ronneberger2015unet}
O.~Ronneberger, P.~Fischer, and T.~Brox.
\newblock {U-Net}: {C}onvolutional networks for biomedical image segmentation.
\newblock In {\em ICMICCI}, 2015.

\bibitem{AziziFar2018}
N.~Azizi, H.~Farazi, and S.~Behnke.
\newblock Location dependency in video prediction.
\newblock In {\em ICANN}, 2018.

\bibitem{Wang2018PredRNNTA}
Y.~Wang, Z.~Gao, M.~Long, J.~Wang, and P.~Yu.
\newblock {PredRNN++}: {T}owards a resolution of the deep-in-time
  spatiotemporal predictive learning.
\newblock In {\em ICML}, 2018.

\bibitem{Wagner2017}
J.~Wagner, V.~Fischer, M.~Herman, and S.~Behnke.
\newblock Learning semantic prediction using pretrained deep feedforward
  networks.
\newblock In {\em ESANN}, 2017.

\bibitem{bei2021learning}
X.~Bei, Y.~Yang, and S.~Soatto.
\newblock Learning semantic-aware dynamics for video prediction.
\newblock In {\em CVPR}, 2021.

\bibitem{luc2017predicting}
P.~Luc, N.~Neverova, C.~Couprie, J.~Verbeek, and Y.~LeCun.
\newblock Predicting deeper into the future of semantic segmentation.
\newblock In {\em ICCV}, 2017.

\bibitem{farazi2019}
H.~Farazi and S.~Behnke.
\newblock Frequency domain transformer networks for video prediction.
\newblock In {\em ESANN}, 2019.

\end{thebibliography}
\end{document}